\documentclass[letterpaper, 10 pt, conference]{ieeeconf}  

\IEEEoverridecommandlockouts                              

\usepackage{cite}
\usepackage{amsmath,amssymb,amsfonts}
\usepackage{algorithmic}
\usepackage{graphicx}
\usepackage[dvipsnames]{xcolor}

\title{\LARGE \bf
 Robot Inner Attention Modeling for Task-Adaptive Teaming of Heterogeneous Multi Robots
}

\author{Chao Huang$^{1}$ and Rui Liu$^{1*}$
\thanks{$^{1}$ is with The Cognitive Robotics and AI Lab (CRAI), College of Aeronautics and Engineering, Kent State University, Kent, OH 44240, USA.}%

\thanks{$^{*}$ Rui Liu is the corresponding author, email: ruiliu.robotics@gmail.com.}
}

\begin{document}

\maketitle
\thispagestyle{empty}
\pagestyle{empty}

\begin{abstract}
Attracted by team scale and function diversity, a heterogeneous multi-robot system (HMRS), where multiple robots with different functions and numbers are coordinated to perform tasks, has been widely used for complex and large-scale scenarios, including disaster search and rescue, site surveillance, and social security. However, due to the variety of the task requirements, it is challenging to accurately compose a robot team with appropriate sizes and functions to dynamically satisfy task needs while limiting the robot resource cost to a low level. To solve this problem, in this paper, a novel adaptive cooperation method, inner attention (\textbf{\textit{innerATT}}), is developed to flexibly team heterogeneous robots to execute tasks as task types and environment change. \textbf{\textit{innerATT}} is designed by integrating a novel attention mechanism into a multi-agent actor-critic reinforcement learning architecture. With an attention mechanism, robot capability will be analyzed to flexibly form teams to meet task requirements. Scenarios with different task variety ("Single Task", "Double Task", and "Mixed Task") were designed. The effectiveness of the \textbf{\textit{innerATT}} was validated by its accuracy in flexible cooperation.
\end{abstract}


\section{Introduction}

 \begin{figure}
\includegraphics [width=1.0 \columnwidth ]{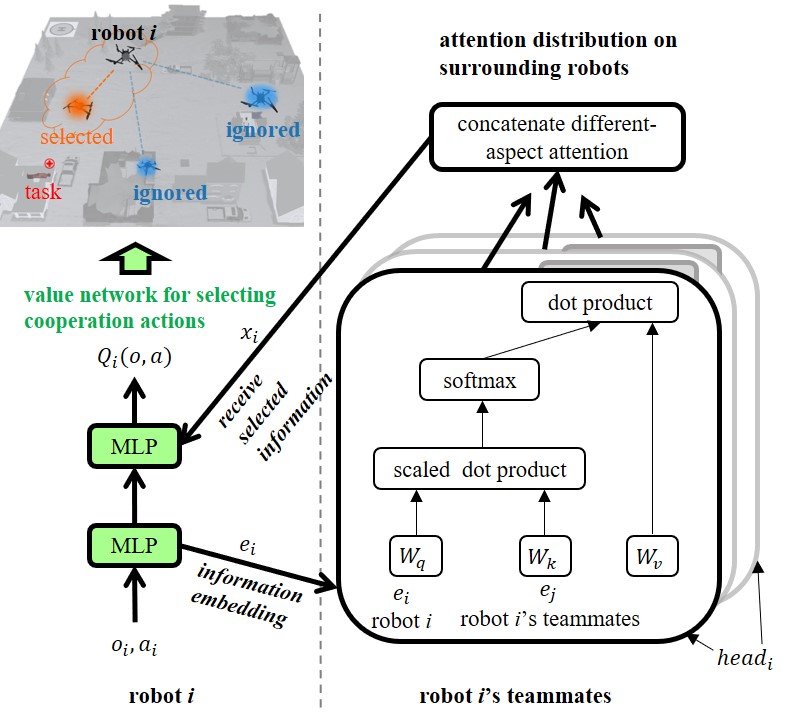}
\caption{The architecture of \textbf{\textit{innerATT}}. The inner attention mechanism determines the attention weights between robots. As the left figure shows, the input is a robot’s observations, status, and actions, and cooperability related information from the robot's teammates. The output is the $Q$-value network for cooperation strategy selection, in which a robot pays different-level attention to others to form a team for a given task. In the right figure, multiple attention heads are used to evaluate different aspects of the cooperability between a robot and its potential teammates.}
\end{figure}

A heterogeneous multi-robot system (HMRS) consists of multiple robots with various shapes, sizes, and capabilities to perform tasks together. With the advantages of functional diversity, team scalability, and control robustness, HMRS has been widely used for large-scale tasks. For example, in natural disaster search and rescue \cite{7485707,8319200,2937074}, a larger area is searched, and multiple victims are rescued because robots in HMRS can operate in parallel. In traffic condition control, such as traffic flow control, public transportation scheduling \cite{alotaibi2016multi,digani2013towards,digani2014hierarchical}, it is cost-effective to deploy multiple robots with particular capabilities to form a strong team instead of deploying one robot with numerous functions to perform the entire mission. As for area coverage and search with complex tasks \cite{broecker2015hybrid,kolling2008multi,easton2005coverage1}, HMRS can resolve task complexity by allocating the mission across robot team members.  

However, influenced by task varieties, teaming of heterogeneous robots in the real world is challenging.
First, Due to the changing of task requirements across different environments, it is difficult to assess assistance type and scale \cite{zhu2013dynamic, zhu2010improved}, and it is also challenging to accurately map the robot team capability to various task needs \cite{vergnano2012modeling}. Second, real-world factors, such as motor degradation, sensor failure, and robot working status, influence robot availability to deliver assistance \cite{prorok2016fast,saribatur2019finding}. Faulty robots in the robot team may share incorrect information with other members of the team leading the whole team unqualified to the assigned task. The unpredictable nature and negative impacts of these real-world faults limit the number of qualified team members, making it challenging to deploy a qualified robot team with expected capability and team size to satisfy task needs.
Third, influenced by both the dynamics in task needs, robot availability, and environment constraints, it is challenging to accurately map robot team capability to task needs \cite{vergnano2012modeling}. 
Assistance needs are dynamic as tasks vary; obstacles and weather conditions influence the time and feasibility of robot participation in assistance; available robots are with different distances to the requested location.
The above task variety impedes an accurate alignment between robot capability and task needs, limiting the deployment of HMRS in the real world. 
Therefore, there is an urgent need to flexibly compose heterogeneous robot teams for mission deployments. In \cite{atay2006mixed, darrah2005multiple, mosteo2006simulated, juedes2004heuristic, kmiecik2010task}, human pre-defined robot utility functions, including teaming strategies, time efficiency, and robot resource (robot types, numbers, and heading speed) consumption, were used to flexibly team robots. However, the pre-optimized allocation strategies ignore the general relation between teaming and task needs. When task requirements change, which is common in real-world situations, predefined teaming strategies cannot satisfy the assistance needs along the mission timeline. To overcome the above-mentioned limits, in \cite{iijima2017adaptive, d2012re, elfakharany2020towards, fan2018fully, noureddine2017multi, luo2019multi}, a deep reinforcement learning algorithm was used to dynamically model the highly nonlinear relation between situation constraints and task needs to flexibly team robots together. Although reinforcement learning methods can perform real-time teaming guidance by considering key elements related to robot status, task type and location, these methods cannot resist robot faults, which will mislead the teaming strategies and cause mission failures. To reduce the influence of robot failures on HMRS performance, research of self-healing was conducted. \cite{zhang2007self, liu2015gradient} investigated methods for mobile robot networks to maintain the logical and physical topology of the network when robots fail and must be replaced within a formation. However, these researches mainly focused on replacing broken robots, which ignores partial failures such as direction deviation and slowed speed that are likely to be encountered in real-world deployments. Recently, \cite{mathews2017mergeable, mathews2019supervised, pelc2005broadcasting, saulnier2017resilient, liu2019trust} limited the negative influence of robot failures on team performance by restricting unreliable information sharing among robots. However, the influence limits were predefined so that they cannot adapt to real-world dynamic situations where the robot performance and task requirement change all the time.

This paper addresses above issues by designing a novel flexible teaming method, inner attention (\textbf{\textit{innerATT}}), which is developed by integrating a novel attention mechanism into a multi-agent actor-critic reinforcement learning architecture, as shown in Figure 1. \textbf{\textit{innerATT}} enables a robot to pay attention to the communication with its available teammates and capture the cooperation-related factors during teaming; the robots flexibly select cooperators to form a team to adapt to environment dynamics. The attention mechanism in \textbf{\textit{innerATT}} is automatically obtained during deployment training. This paper mainly has three contributions:
\begin{enumerate}
\item A novel multi-robot teaming method, \textbf{\textit{innerATT}}, has been developed to guide the flexible cooperation among heterogeneous robots as the task complexity varies in target number, target type, and robot work status.
\item A theoretical analysis has been conducted to validate the robustness of \textbf{\textit{innerATT}} in guiding flexible cooperation, providing a theoretical foundation for implementing \textbf{\textit{innerATT}} in general disturbance-involved multi-robot teaming in future similar research.
\item A deep reinforcement learning based simulation framework, which integrates the simulation platform of multi-agent particle environment, the multi-agent deep reinforcement learning algorithms, and robot models, has been developed to provide standard pipelines for simulating the flexible robot teaming.
\end{enumerate}

\section{Inner Attention Supported Adaptive Cooperation}

With the \textbf{\textit{innerATT}} supported by the inner attention mechanism, a robot assesses the function compensation and cooperation likelihood of its surrounding teammates, estimating a teaming plan based on the available robot sources. As shown in Figure 1, given the inputs are robots' motion status and observations of the surrounding environment, \textbf{\textit{innerATT}} automatically determines the amount of attention paid to different robots to form a mission team.

\subsection{Robot Inner Attention for Team Adaptability Modeling}
The basic framework of robot teaming is established based on a multi-agent actor-critic deep reinforcement learning (MAAC) algorithm defined by robot number, \begin{math} N \end{math};  motion state space, \begin{math} S \end{math}; robot action sets, \begin{math} A = \{A_1, ... A_N\} \end{math}; transition probability function over the next possible states, \begin{math} T\end{math}: \begin{math} S \times A_1 \times ... \times A_N \rightarrow P(S)\end{math}; a set of observations for all robots, \begin{math} O = \{O_1, ... O_N\} \end{math}; and reward function for each robot \begin{math} R_i\end{math}: \begin{math}S \times A_1 \times ... \times A_N \rightarrow R \end{math}. By using extended actor-critic reinforcement learning to guide robot cooperation, each robot learns an individual policy function, \begin{math} \pi_i\end{math}: \begin{math}O_i \rightarrow P(A_i) \end{math}, which is a probability distribution on potential cooperation actions. The goal of multi-agent reinforcement learning is to learn an optimal cooperation strategy for each robot that can maximize their expected discounted returns: \begin{equation} J_i(\pi_i) = E_{a_* \sim \pi_*; s \sim T}[\sum_{t=0}^{\infty}\gamma^tr_{it}(s_t, a_{1t}, ...,a_{Nt})], \end{equation} where * represent  \begin{math} \{1, ... N\} \end{math}; \begin{math} \gamma \in [0, 1] \end{math} is the discount factor that determines the degree to which the cooperation policy favors immediate reward over long-term gain.

In the extended actor-critic framework consisting of centralized training with decentralized execution, to calculate the \begin{math} Q \end{math}-value function \begin{math} Q_i(o,a) \end{math} for the robot \begin{math} i \end{math}, the critic receives the observations, \begin{math} o = (o_1, ..., o_N) \end{math}, and actions, \begin{math} a = (a_1, ..., a_N) \end{math}, for all robots, which will take redundant information into account. With an attention mechanism, each robot actively explores cooperation-relevant information from its surrounding robots to assess their compatibility and potential collaboration, and therefore selectively choose robots to team together. 

To generate the attention weights, the embedded information is fed into the \textbf{\textit{innerATT}} to get the \begin{math} Q \end{math}-value function \begin{math} Q_i(o; a) \end{math} for robot \begin{math} i \end{math}, which is a function: \begin{equation} Q_i(o; a) = w^{2T}\sigma(w^1, <e_i, x_i>), \end{equation} where \begin{math} \sigma \end{math} is rectified linear units (ReLU), \begin{math} w^1 \end{math} and \begin{math} w^2 \end{math} are the parameters of critics. The inner attention mechanism has shared query (\begin{math} w_q \end{math}), key (\begin{math} w_k \end{math}), and value (\begin{math} w_v \end{math}) matrixes. Each robot's embedding \begin{math} e_i \end{math} can be linearly transformed into \begin{math} q_i\end{math}, \begin{math}k_i \end{math}, and \begin{math} v_i \end{math} separately. The contribution from other robots, \begin{math} x_i \end{math}, is a weighted sum of other robots’ value: \begin{equation} x_i = \sum_{j \neq i}\alpha_{ij}v_j = \sum_{j \neq i}\alpha_{ij}\sigma(v_j). \end{equation} The attention weight \begin{math} \alpha_{ij} \end{math} compares the similarity between \begin{math} k_j \end{math} and \begin{math} q_i \end{math}, and the similarity value can be obtained from a softmax function: \begin{equation} \alpha_{ij} =\frac{S_{ij}}{\sum_{k=1}^N{S_{ik}}}= \frac{e_jw_k^Tw_qe_i}{\sum_{k=1}^Ne_kw_k^Tw_qe_i}.\end{equation} 
To better analyze the effectiveness of the \textbf{\textit{innerATT}} method, a baseline method without the inner attention mechanism has also been designed. In the baseline method, the attention weights \begin{math} \alpha \end{math} are simply fixed to \begin{math} \frac{1}{(N-1)} \end{math}. Given that only the values of attention weights are changed to a fixed value, both \textbf{\textit{innerATT}} and baseline methods are implemented with an approximately equal number of parameters.

\subsection{Theoretical analysis of \textbf{\textit{innerATT}}'s robustness}
To simply explain whether inner attention mechanism works in supporting flexible teaming, the output of the $Q$-value neural network with inner attention mechanism, when the input is \begin{math} x \end{math}, can be written as: \begin{equation} f(x) = w^{2T}\sigma(w^1, x), x = <e_i, x_i>, \end{equation} the robots can be more robust to other robots' failure or a sensor failure \cite{hsieh2019robustness}. Consider that a small perturbation is added to a particular robot \begin{math} j \end{math}'s embedding, such that \begin{math} e_j \end{math} is changed to \begin{math} e_j + \bigtriangleup e \end{math} while all the other robots' embeddings remain unchanged. How much will this perturbation affect the attention weights \begin{math} \alpha_{ij} \end{math}? For a particular \begin{math} i (i \neq j) \end{math}, the \begin{equation} S_{ij} = e_jw_k^Tw_qe_i \end{equation} is only changed by one term since: \begin{equation} S'_{ij} = \begin{cases} S_{ij} + \bigtriangleup ew_k^Tw_qe_i, & if (i \neq j) \\ S_{ij}, & otherwise \end{cases}, \end{equation} where \begin{math} S'_{ij} \end{math} denotes the value after the perturbation. Therefore, with the perturbed input, each set of \begin{math} \{S_{ij}\}_{j=1}^N \end{math} will only have one term being changed. 
For the perturbation part, assume \begin{math} \| \bigtriangleup e \| \leq \delta_1 \end{math} and \begin{math} \| e_i \| \leq \delta_2 \end{math}, then the expected value is \begin{equation} E[S'_{ij} - S_{ij}] \leq \|w_q\|\|w_k\|\delta_1\delta_2.\end{equation} Then, the probability results can be obtained by using Markov inequality: \begin{equation} P(|S'_{ij} - S_{ij}| \geq \varepsilon) \leq \frac{\|w_q\|\|w_k\|\delta_1\delta_2}{\varepsilon}. \end{equation}

Therefore, as the norm of \begin{math} w_q, w_k \end{math} are not too large (usually regularized by \begin{math} L_2 \end{math} during training), there will be a significant amount of \begin{math} i \end{math} such that \begin{math} S'_{ij} \end{math} is perturbed negligibly. Therefore, with the inner attention mechanism, \textbf{\textit{innerATT}} method is robust to the robot unavailability caused by either mission occupation or system failures.

\section{Experiment Settings}

To validate \textbf{\textit{innerATT}}'s effectiveness in improving HMRS adaptability, a task environment with three variety situations ("Single Task", "Double Task", and "Mixed Task") were designed. These scenarios are typical task situations in real-world mission deployment, therefore being used here to evaluate the general effectiveness of the proposed method in mission support.

The environment shown in Figure 2 was implemented based on the open-source multi-agent particle environment (MPE) framework \cite{lowe2017multi}. The size of the artificial environment was set to 2 \begin{math} \times \end{math} 2. The parameters of robots, shown in Table I, were set according to real-world robots. Two types of victims and four types of rescue robots are involved to simulate the rescuing process. For the rescue robots, two of them are food delivery robots providing living supplies such as food and water; one is a navigation robot providing victims with useful location information, directing them to safer places. The remaining robots are medical assistance robots to provide medical treatments to heavily injured victims. For victims, one type is heavily injured, requiring both food and medical assistance for survival, defined as "Task 1"; while another is trapped but in good health, needing food as well as navigation, defined as "Task 2".

\begin{figure}[t!]
  \centering
  \includegraphics [width=0.95\columnwidth ]{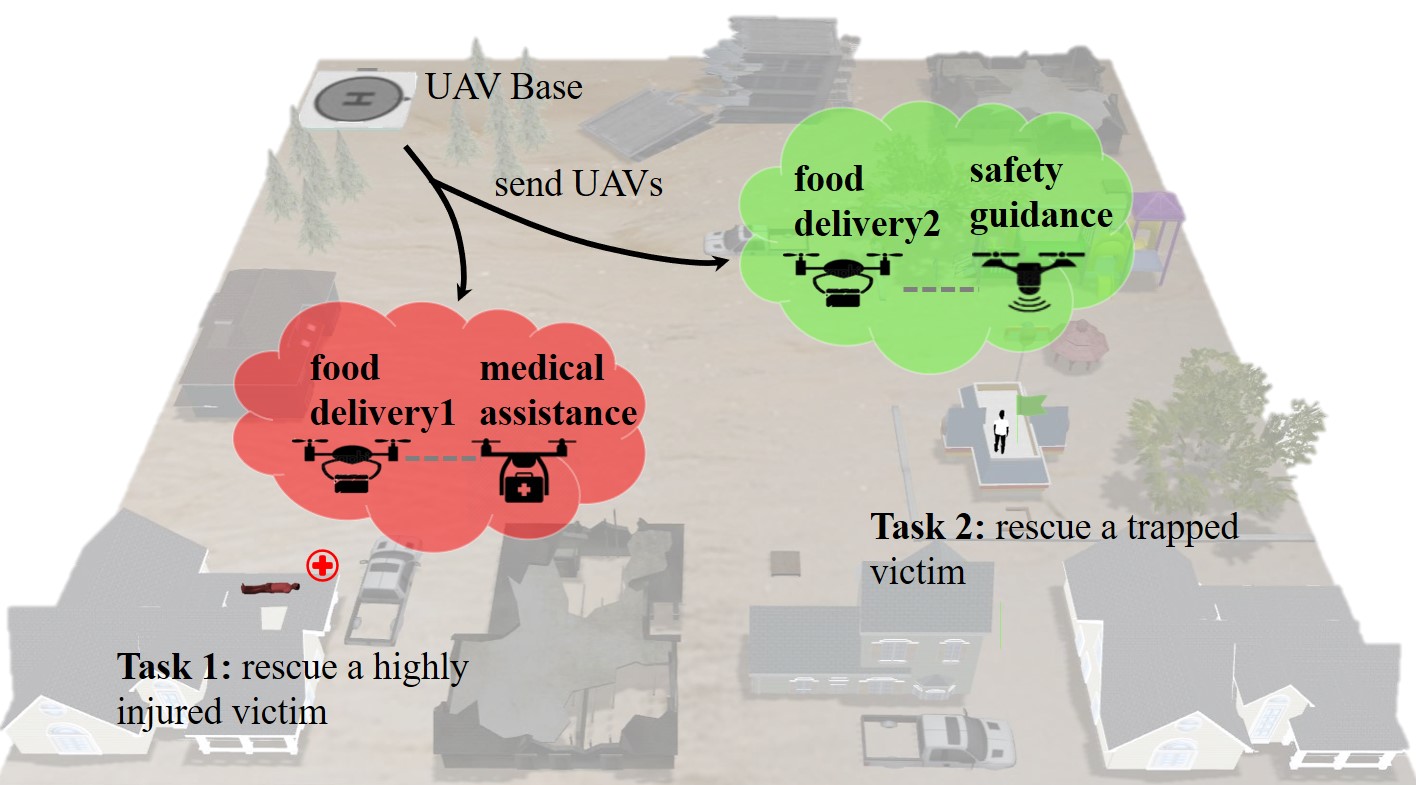}
  \caption{Simulated environment scenario illustration. In the flood disaster, there are trapped victims with different injury levels. For the victims with high injury level (Task 1), they need rescue robots providing them with food, water, and emergency medical treatment; While for the victims with low injury level (Task 2), they will need other kinds rescue robots providing food, water, and useful information to guide them to safer places. The main robot team is expected to split into different sub-teams that can rescue these victims effectively.}
\end{figure}

\begin{table}
\centering
\caption{The configurations of robots}
\begin{IEEEeqnarraybox}[\IEEEeqnarraystrutmode\IEEEeqnarraystrutsizeadd{2pt}{0pt}]{x/r/Vx/r/x/r/x}
\IEEEeqnarraydblrulerowcut\\
&\IEEEeqnarraymulticol{3}{t}{Type}\hfill&\hfill
\hfill\raisebox{0pt}[0pt][0pt]{Speed}\hfill&\hfill
\hfill\raisebox{0pt}[0pt][0pt]{Mass}\hfill&\hfill
\hfill\raisebox{0pt}[0pt][0pt]{Ability}\hfill&\\
\IEEEeqnarrayrulerow\\
&\IEEEeqnarraymulticol{3}{t}{Food Delivery}\hfill&\hfill
\raisebox{0pt}[0pt][0pt]{1.0 m/s}&\hfill
\raisebox{0pt}[0pt][0pt]{1.0 kg}&\hfill
\hfill\raisebox{0pt}[0pt][0pt]{Food}\hfill&\\
&\IEEEeqnarraymulticol{3}{t}{Navigation}\hfill&\hfill
\raisebox{0pt}[0pt][0pt]{1.5 m/s}&\hfill
\raisebox{0pt}[0pt][0pt]{0.5 kg}&\hfill
\hfill\raisebox{0pt}[0pt][0pt]{Information}\hfill&\\
&\IEEEeqnarraymulticol{3}{t}{Medical Assistance}\hfill&\hfill
\raisebox{0pt}[0pt][0pt]{1.5 m/s}&\hfill
\raisebox{0pt}[0pt][0pt]{0.5 kg}&\hfill
\hfill\raisebox{0pt}[0pt][0pt]{Medicine}\hfill&\\
\IEEEeqnarraydblrulerowcut
\end{IEEEeqnarraybox}
\end{table} 

\begin{figure}[t!]
  \centering
 \includegraphics [width=1\columnwidth ]{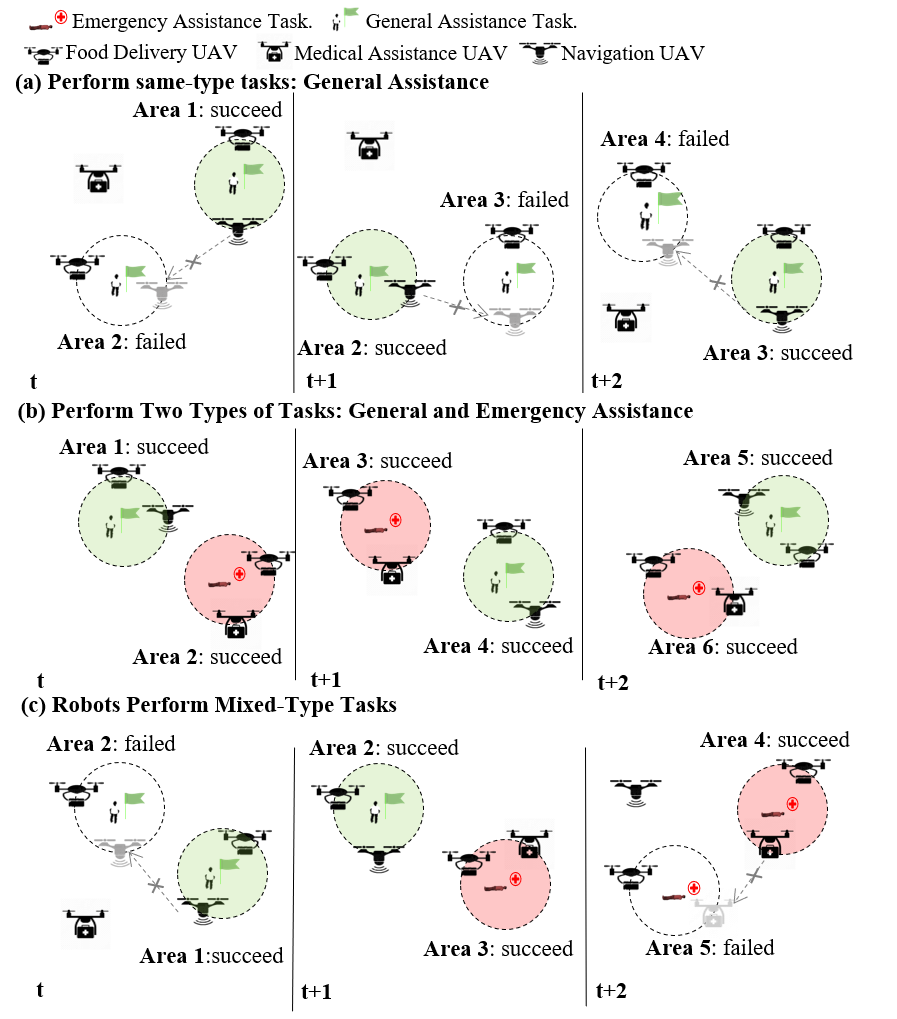}
  \caption{Simulated environment situations illustration. (a) "($S_1$) Single Task" in which only one kind of task popup. (b) "($S_2$) Double Task" in which always popup two kind of tasks. Robot team will always deal with Task1 and Task2. (c) "($S_3$) Mixed Task" which is the combination of situation one and two.}
  \label{env}
\end{figure}

To analyze the effectiveness of the \textbf{\textit{innerATT}} method, three situations - from simple to complex - for each scenario were designed, shown in Figure 3. In "$(S_1)$ Single Task" situation, only one task, "Task 1" or "Task 2", appeared at random locations. In "$(S_2)$ Double Task", "Task 1" and "Task 2" will always present together in each time at random locations. In "$(S_3)$ Mixed Task" is a combination of $S_1$ and $S_2$, that is an unpredictable number of tasks appear at a random location each time. In addition, two deep reinforcement learning algorithms with (simplified as PPO) and without (simplified as TD) proximal policy optimization have also been used. In the method without inner attention, the attention weights \begin{math} \alpha \end{math} are simply fixed to \begin{math} \frac{1}{(N-1)} \end{math}. Given that only the values of attention weights are changed to a fixed value, both \textbf{\textit{innerATT}} and methods without inner attention are implemented with an approximately equal number of parameters.

As for the training procedure, the extended actor-critic method for maximum entropy reinforcement learning was used in the training progress of 25,000 episodes. There were 12 threads to process training data in parallel and a replay buffer to store experience tuples of \begin{math} (o_t, a_t, r_t, o_{t+1}) \end{math} for each time step.  In detail, sample 1024 tuples from the replay buffer and update the parameters of the \begin{math} Q \end{math}-function and the policy objective through policy gradients. Adam optimizer was used, and the initial learning rate was set as 0.001 and the discount factor \begin{math} \gamma \end{math} was 0.99. The embedded information function used a hidden dimension of 128, and four attention heads were used in the inner attention mechanism. 

\section{Results}
\subsection{Adapting to Task Varieties}

\begin{figure}[t!]
  \centering
 \includegraphics [width=1\columnwidth ]{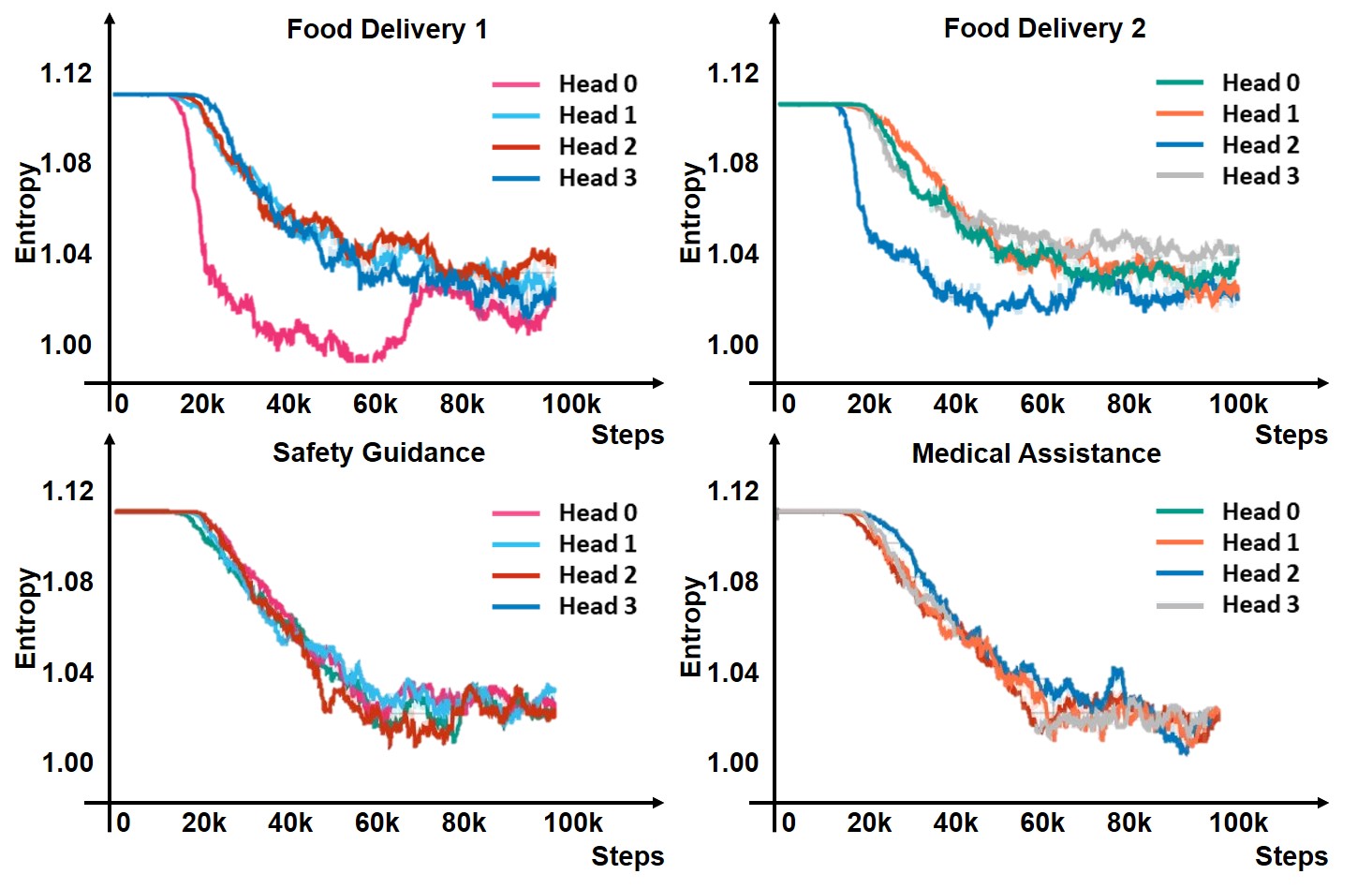}
  \caption{Attention entropy of each attention head during the training phase for the robots in the multi-robot cooperation environment. A lower entropy value indicates that the robots have learned to selectively pay attention to another specific team member.}

\end{figure}

\begin{figure*}[t!]
  \centering
 \includegraphics [width=2.04\columnwidth ]{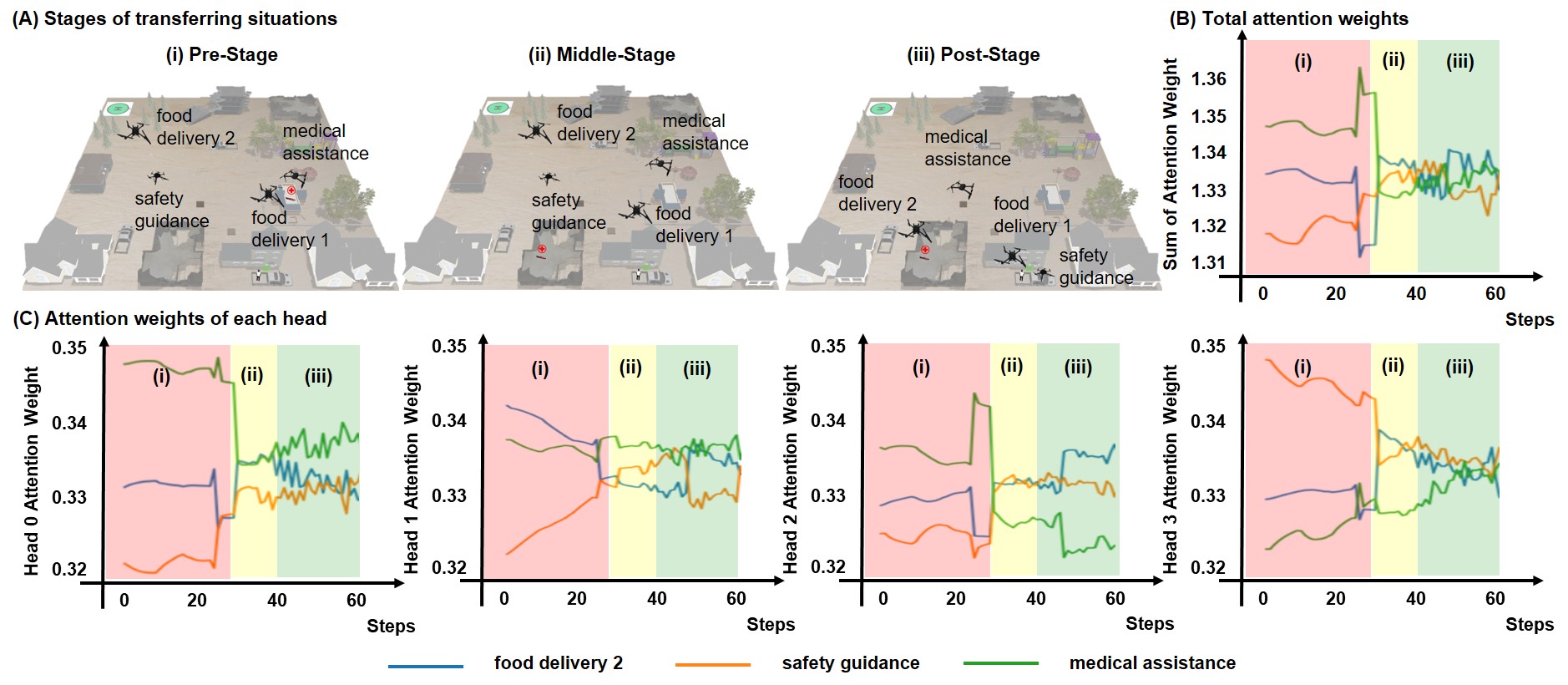}
  \caption{Relationships between food delivery 1 robot's behavior and its inner attention weights in adaptive teaming. (A) three stages of food delivery 1 robot's flexible teaming. In pre-stage (i), food delivery 1 robot is cooperating with medical assistance robot. In middle-stage (ii), food delivery 1 robot is changing its behavior based on inner attention mechanism. In post-stage (iii), food delivery 1 robot is cooperating with a navigation robot. (B) food delivery 1 robot's total attention weight paid to other robots. (C) food delivery 1 robot's attention weights obtained from each attention head.}
\end{figure*}

In the typical "task variety" scenario, robots' flexibility, the cooperation rate between food delivery robots and other rescue robots, was calculated in a period of time (80 episodes) by using the following formulation: \begin{equation} rate_{ij} = \frac{Num_{ij}}{\sum_{k=1}^NNum_{ik}}, \end{equation} where \begin{math} \sum_{k=1}^NNum_{ik} \end{math} is the total number of victims rescued by robot \begin{math}i\end{math}; \begin{math} Num_{ij} \end{math} is the total number of victims rescued by the cooperation of robot \begin{math}i\end{math} and robot \begin{math}j\end{math}. The results are shown in Table II. In Task 1, the cooperation rates of food delivery robots trained by \textbf{\textit{innerATT}} are 0.52 and 0.48 respectively, which is similar to the uniform distribution with 95\% confidence; while the cooperation rates of food delivery robots trained by baseline method are 0.90 and 0.10, which doesn't have enough evidence to prove that it is similar to the uniform distribution. Similar results have been shown for task 2, that the robots trained by \textbf{\textit{innerATT}} are more flexible than those trained by the baseline method. As suspected, the baseline model's critics use all information non-selectively, while \textbf{\textit{innerATT}} can learn which robot to pay more attention to through the inner attention mechanism. Thus, \textbf{\textit{innerATT}} method is more flexible and sensitive to dynamically changed tasks. Besides that, Figure 4 demonstrates the effect of the attention head on the robot during the training process by showing the entropy of the attention weights for each robot. From the results shown in Figure 4, the entropy of all robot attention heads is continually decreasing to 1.02 around, which indicates that \textbf{\textit{innerATT}} can train the robots to selectively pay attention to a specific team member through the inner attention mechanism.

\begin{table}
\centering
\caption{UAVs participate rate comparison}
\begin{IEEEeqnarraybox}[\IEEEeqnarraystrutmode\IEEEeqnarraystrutsizeadd{2pt}{0pt}]{x/r/vx/r/V/r/x/r/x/r/x}
\IEEEeqnarraydblrulerowcut\\
&\IEEEeqnarraymulticol{3}{t}{}&&&\hfill
\raisebox{0pt}[0pt][0pt]{UAV food}&&\hfill
\raisebox{0pt}[0pt][0pt]{UAV food}&&\hfill
\chi_1^2\hfill&\\
&\IEEEeqnarraymulticol{3}{t}{}&&&\hfill
\raisebox{0pt}[0pt][0pt]{delivery1}&&\hfill
\raisebox{0pt}[0pt][0pt]{delivery2}&&\hfill
(a=0.05)\hfill&\\
\IEEEeqnarrayrulerow\\
&&&&\hfill \raisebox{-0.5pt}[0pt][0pt]{TD-innerATT} \hfill&&\hfill
\raisebox{-0.5pt}[0pt][0pt]{0.47}\hfill&&\hfill
\raisebox{-0.5pt}[0pt][0pt]{0.53}\hfill&&\hfill
\raisebox{-0.5pt}[0pt][0pt]{0.36 $<$ 3.84}\hfill&
\IEEEeqnarraystrutsizeadd{0pt}{2pt}\\
&\hfill\raisebox{-2.5pt}[0pt][0pt]{Task1}\hfill&&
\IEEEeqnarraymulticol{9}{h}{}%
\IEEEeqnarraystrutsize{0pt}{0pt}\\
&&&&\hfill\raisebox{-0.5pt}[0pt][0pt]{TD}\hfill&&\hfill
\raisebox{-0.5pt}[0pt][0pt]{0.82}\hfill&&\hfill
\raisebox{-0.5pt}[0pt][0pt]{0.18}\hfill&&\hfill
\raisebox{-0.5pt}[0pt][0pt]{81.9 $>$ 3.84}\hfill&
\IEEEeqnarraystrutsizeadd{0pt}{2pt}\\
\IEEEeqnarrayrulerow\\
&&&&\hfill\raisebox{-0.5pt}[0pt][0pt]{TD-innerATT}\hfill&&\hfill
\raisebox{-0.5pt}[0pt][0pt]{0.56}\hfill&&\hfill
\raisebox{-0.5pt}[0pt][0pt]{0.44}\hfill&&\hfill
\raisebox{-0.5pt}[0pt][0pt]{1.44 $<$ 3.84}\hfill&
\IEEEeqnarraystrutsizeadd{0pt}{2pt}\\
&\hfill\raisebox{-3pt}[0pt][0pt]{Task2}\hfill&&
\IEEEeqnarraymulticol{9}{h}{}%
\IEEEeqnarraystrutsize{0pt}{0pt}\\
&&&&\hfill\raisebox{-0.5pt}[0pt][0pt]{TD}\hfill&&\hfill
\raisebox{-0.5pt}[0pt][0pt]{0.18}\hfill&&\hfill
\raisebox{-0.5pt}[0pt][0pt]{0.82}\hfill&&\hfill
\raisebox{-0.5pt}[0pt][0pt]{81.9 $>$ 3.84}\hfill&
\IEEEeqnarraystrutsizeadd{0pt}{2pt}\\
\IEEEeqnarrayrulerow\\
&&&&\hfill \raisebox{-0.5pt}[0pt][0pt]{PPO-innerATT} \hfill&&\hfill
\raisebox{-0.5pt}[0pt][0pt]{0.48}\hfill&&\hfill
\raisebox{-0.5pt}[0pt][0pt]{0.52}\hfill&&\hfill
\raisebox{-0.5pt}[0pt][0pt]{0.16 $<$ 3.84}\hfill&
\IEEEeqnarraystrutsizeadd{0pt}{2pt}\\
&\hfill\raisebox{-2.5pt}[0pt][0pt]{Task1}\hfill&&
\IEEEeqnarraymulticol{9}{h}{}%
\IEEEeqnarraystrutsize{0pt}{0pt}\\
&&&&\hfill\raisebox{-0.5pt}[0pt][0pt]{PPO}\hfill&&\hfill
\raisebox{-0.5pt}[0pt][0pt]{0.32}\hfill&&\hfill
\raisebox{-0.5pt}[0pt][0pt]{0.68}\hfill&&\hfill
\raisebox{-0.5pt}[0pt][0pt]{25.9 $>$ 3.84}\hfill&
\IEEEeqnarraystrutsizeadd{0pt}{2pt}\\
\IEEEeqnarrayrulerow\\
&&&&\hfill\raisebox{-0.5pt}[0pt][0pt]{PPO-innerATT}\hfill&&\hfill
\raisebox{-0.5pt}[0pt][0pt]{0.45}\hfill&&\hfill
\raisebox{-0.5pt}[0pt][0pt]{0.55}\hfill&&\hfill
\raisebox{-0.5pt}[0pt][0pt]{1.00 $<$ 3.84}\hfill&
\IEEEeqnarraystrutsizeadd{0pt}{2pt}\\
&\hfill\raisebox{-3pt}[0pt][0pt]{Task2}\hfill&&
\IEEEeqnarraymulticol{9}{h}{}%
\IEEEeqnarraystrutsize{0pt}{0pt}\\
&&&&\hfill\raisebox{-0.5pt}[0pt][0pt]{PPO}\hfill&&\hfill
\raisebox{-0.5pt}[0pt][0pt]{0.73}\hfill&&\hfill
\raisebox{-0.5pt}[0pt][0pt]{0.27}\hfill&&\hfill
\raisebox{-0.5pt}[0pt][0pt]{42.3 $>$ 3.84}\hfill&
\IEEEeqnarraystrutsizeadd{0pt}{2pt}\\
\IEEEeqnarraydblrulerowcut\\
\end{IEEEeqnarraybox}
\end{table} 

To further prove that the inner attention mechanism is beneficial to robot's flexible adaptation to different tasks, the relationship between robot behaviors and their inner attention weights was analyzed to illustrate attention supports in adjusting robot behaviors for flexible teaming. Figure 5 (A) is an illustration of a specific scenario occurring during the experiment. In the pre-stage, food delivery 1 robot is firstly cooperating with medical assistance robot to rescue the heavily injured victim (Task 1). At this moment, food delivery 1 robot needs to pay more attention to medical assistance robot. After finishing Task 1, in the middle-stage and post-stage, it will change to cooperate with a navigation robot to rescue the trapped victim in good health (Task 2). At this time, food delivery 1 robot needs to pay more attention to navigation robot. Figure 5 (B) is the curves of food delivery 1 robot's total attention weights over the other three robots. In the pre-stage, the curve of total attention weights paid on medical assistance robot has the highest values, which supports the food delivery 1 robot to selectively cooperate with medical assistance robot. In the middle-stage and post-stage, the curves of total attention weights paid on medical assistance robot and navigation robot are decreasing and increasing separately, which supports food delivery 1 robot to transfer its attention from medical assistance robot to navigation robot. Therefore, the inner attention mechanism can support robot flexible teaming behaviors to different tasks. Figure 5 (C) are the curves of food delivery 1 robot's attention weights, generated by each attention head, over other rescue robots. 

\section{Conclusion and Future Work}
This paper developed a novel inner attention model, \textbf{\textit{innerATT}}, to enable multi heterogeneous robots to cooperate flexibly according to task needs. With scenarios of different task varieties, including "a single task, double task, and dynamically mixed tasks", the effectiveness of the \textbf{\textit{innerATT}} model for guiding flexible teaming has been validated. This model essentially addressed the question of allocating limited available robot sources into dynamic task situations. This theoretical model can also be extended to guide flexible teaming between the ground and aerial vehicles, and even the teaming between vehicles and human units. Therefore, this attention-based flexible teaming model bears a huge potential for real-world multi-robot implementations, from disaster research, to wildlife protection, and to airport traffic control.

Notes that, in this work, the primary focus is validating the feasibility of using attention for flexible heterogeneous teaming. The simulated environment is different from the real-world environment, so the model trained in our research cannot achieve the same performance in real-world applications; but the trained model will be a helpful initial learner for further training in the real-world. In the future, the research of robot behavior understanding and human trust modeling will be an option to improve the performance of HMRS in the real world.

\addtolength{\textheight}{0cm}   






\begin{thebibliography}{99}
\bibitem{7485707}A. Matos, A. Martins, A. Dias, B. Ferreira, J. M. Almeida, H. Ferreira, G. Amaral, A. Figueiredo, R. Almeida and F. Silva, "Multiple robot operations for maritime search and rescue in euRathlon 2015 competition," OCEANS 2016-Shanghai, pp. 1-7, 2016.

\bibitem{8319200}C. Mouradian, S. Yangui and R. H. Glitho, "Robots as-a-service in cloud computing: search and rescue in large-scale disasters case study," 2018 15th IEEE Annual Consumer Communications and Networking Conference (CCNC), pp. 1-7, 2018.

\bibitem{2937074}Z. Beck, Teacy, N. R. Jennings and A. C. Rogers, "Online planning for collaborative search and rescue by heterogeneous robot teams," Association of Computing Machinery, 2016.

\bibitem{alotaibi2016multi}E. T. S. Alotaibi, H. Al-Rawi, "Multi-robot path-planning problem for a heavy traffic control application: A survey," International Journal of Advanced Computer Science and Applications, vol. 7, no. 6, pp. 10, 2016.

\bibitem{digani2013towards}V. Digani, L. Sabattini, C. Secchi and C. Fantuzzi, "Towards decentralized coordination of multi robot systems in industrial environments: A hierarchical traffic control strategy," 2013 IEEE 9th International Conference on Intelligent Computer Communication and Processing (ICCP), pp. 209-215, 2013.

\bibitem{digani2014hierarchical}V. Digani, L. Sabattini, C. Secchi and C. Fantuzzi, "Hierarchical traffic control for partially decentralized coordination of multi agv systems in industrial environments," IEEE International Conference on Robotics and Automation, pp. 6144-6149, 2014.

\bibitem{broecker2015hybrid}B. Broecker, I. Caliskanelli, K. Tuyls, E. I. Sklar and D. Hennes, "Hybrid insect-inspired multi-robot coverage in complex environments." Conference Towards Autonomous Robotic Systems, pp. 56-68, 2015.

\bibitem{kolling2008multi}A. Kolling and S. Carpin, "Multi-robot surveillance: an improved algorithm for the graph-clear problem," IEEE International Conference on Robotics and Automation, pp. 2360-2365, 2008.

\bibitem{easton2005coverage1}K. Easton and J. Burdick,  "A coverage algorithm for multi-robot boundary inspection," IEEE International Conference on Robotics and Automation, pp. 727-734, 2005.

\bibitem{zhu2013dynamic}D. Q. Zhu, H. Huang and S. X. Yang, "Dynamic task assignment and path planning of multi-AUV system based on an improved self-organizing map and velocity synthesis method in three-dimensional underwater workspace," IEEE Transactions on Cybernetics, vol. 43, no. 2,  pp. 504-514, 2013.

\bibitem{zhu2010improved}A. M. Zhu and S. X. Yang, "An improved SOM-based approach to dynamic task assignment of multi-robot," World Congress on Intelligent Control and Automation, pp. 2168-2173, 2010.



\bibitem{prorok2016fast}A. Prorok, M. A. Hsieh and V. Kumar, "Fast redistribution of a swarm of heterogeneous robots," Proceedings of the 9th EAI International Conference on Bio-inspired Information and Communications Technologies (formerly BIONETICS), pp. 249-255, 2016.

\bibitem{saribatur2019finding}Z. G. Saribatur, V. Patoglu and E. Erdem, "Finding optimal feasible global plans for multiple teams of heterogeneous robots using hybrid reasoning: an application to cognitive factories," Autonomous Robots, vol. 43, no. 1, pp. 213-238, 2019.

\bibitem{vergnano2012modeling}A. Vergnano, C. Thorstensson, B. Lennartson, P. Falkman, M. Pellicciari, F. Leali and S. Biller, "Modeling and optimization of energy consumption in cooperative multi-robot systems," IEEE Transactions on Automation Science and Engineering, vol. 9, no. 2, pp. 423-428, 2012.

\bibitem{atay2006mixed}N. Atay and B. Bayazit, "Mixed-integer linear programming solution to multi-robot task allocation problem," 2006.

\bibitem{darrah2005multiple}M. Darrah, W. Niland and B. Stolarik, "Multiple UAV dynamic task allocation using mixed integer linear programming in a SEAD mission," Infotech at Aerospace, pp. 7165, 2005.

\bibitem{mosteo2006simulated}A. R. Mosteo and L. Montano, "Simulated annealing for multi-robot hierarchical task allocation with flexible constraints and objective functions," Workshop on Network Robot Systems: Toward Intelligent Robotic Systems Integrated with Environments, 2006.

\bibitem{juedes2004heuristic}D. Juedes, F. Drews, L. Welch and D. Fleeman, "Heuristic resource allocation algorithms for maximizing allowable workload in dynamic, distributed real-time systems," International Parallel and Distributed Processing Symposium, pp. 117, 2004.

\bibitem{kmiecik2010task}W. Kmiecik, M. Wojcikowski, L. Koszalka and A. Kasprzak, "Task allocation in mesh connected processors with local search meta-heuristic algorithms," Asian Conference on Intelligent Information and Database Systems, pp. 215-224, 2010.

\bibitem{iijima2017adaptive}N. Iijima, A. Sugiyama, M. Hayano and T. Sugawara, "Adaptive task allocation based on social utility and individual preference in distributed environments," Procedia computer science, vol. 112, pp. 91-98, 2017.

\bibitem{d2012re}D. Lope, Javier, D. Maravall and Y. Quiñonez, "Response threshold models and stochastic learning automata for self-coordination of heterogeneous multi-tasks distribution in multi-robot systems," Robotics and Autonomous Systems, 2012.

\bibitem{elfakharany2020towards}A. Elfakharany, R. Yusof, Z. Ismail, "Towards multi-robot Task Allocation and Navigation using Deep Reinforcement Learning," Journal of Physics: Conference Series, vol. 1447, no. 1, pp. 012045, 2020.

\bibitem{fan2018fully}T. X. Fan, P. X. Long, W. X. Liu, J. Pan, "Fully distributed multi-robot collision avoidance via deep reinforcement learning for safe and efficient navigation in complex scenarios," arXiv preprint arXiv:1808.03841, 2018.

\bibitem{noureddine2017multi}D. B. Noureddine, A. Gharbi and S. B. Ahmed, "Multi-agent Deep Reinforcement Learning for Task Allocation in Dynamic Environment," ICSOFT, pp. 17-26, 2017.

\bibitem{luo2019multi}T. Z. Luo, B. Subagdja, D. Wang and A. Tan, "Multi-Agent Collaborative Exploration through Graph-based Deep Reinforcement Learning," 2019 IEEE International Conference on Agents (ICA), pp. 2-7, 2019.

\bibitem{liu2015gradient}Z. Liu, J. Ju, W. Chen, X. Y. Fu and H. Wang, "A gradient-based self-healing algorithm for mobile robot formation," 2015 IEEE/RSJ International Conference on Intelligent Robots and Systems (IROS), pp. 3395-3400, 2015.

\bibitem{zhang2007self}F. Zhang and W. Chen, "Self-healing for mobile robot networks with motion synchronization," 2007 IEEE/RSJ International Conference on Intelligent Robots and Systems, pp. 3107-3112, 2007.

\bibitem{mathews2017mergeable}N. Mathews, A. L. Christensen, R. O’Grady, F. Mondada and M. Dorigo, "Mergeable nervous systems for robots," Nature communications, vol. 8, no. 1, pages 1-7, 2017.

\bibitem{mathews2019supervised}N. Mathews, A. L. Christensen, A. Stranieri, A. Scheidler and M. Dorigo, "Supervised morphogenesis: Exploiting morphological flexibility of self-assembling multirobot systems through cooperation with aerial robots," Robotics and autonomous systems, vol. 112, pp. 154-167, 2019.

\bibitem{pelc2005broadcasting}A. Pelc and D. Peleg, "Broadcasting with locally bounded byzantine faults." Information Processing Letters, vol. 93, no. 3, pp. 109-115, 2005.

\bibitem{saulnier2017resilient}K. Saulnier, D. Saldana, A. Prorok, G. J. Pappas and V. Kumar, "Resilient flocking for mobile robot teams," IEEE Robotics and Automation letters, vol. 2, no. 2, pp. 1039-1046, 2017.

\bibitem{liu2019trust}R. Liu, F. Jia, W. H. Luo, M. Chandarana, C. J. Nam, M. Lewis and K. Sycara, "Trust-Aware Behavior Reflection for Robot Swarm Self-Healing," Proceedings of the 18th International Conference on Autonomous Agents and MultiAgent Systems, pp. 122-130, 2019.

\bibitem{hsieh2019robustness}Y. L. Hsieh, M. H. Cheng, D. C. Juan, W. Wei, W. L. Hsu and C. J. Hsieh, "On the robustness of self-attentive models," Proceedings of the 57th Annual Meeting of the Association for Computational Linguistics, pp. 1520-1529, 2019.

\bibitem{lowe2017multi}R. Lowe, Y. I. Wu, A. Tamar, J. Harb, O. P. Abbeel and I. Mordatch, "Multi-agent actor-critic for mixed cooperative-competitive environments," Advances in neural information processing systems, pp. 6379-6390, 2017.


\end{thebibliography}
\end{document}